\documentclass[10pt,twocolumn,letterpaper]{article}

\usepackage{wacv}
\usepackage{times}
\usepackage{epsfig}
\usepackage{graphicx}
\usepackage{amsmath}
\usepackage{amssymb}
\usepackage{algorithm2e}
\usepackage{threeparttable}
\usepackage{lipsum}

\def\wacvPaperID{****} 

\wacvfinalcopy 

\ifwacvfinal
\def\assignedStartPage{1} 
\fi

\ifwacvfinal
\usepackage[breaklinks=true,bookmarks=false]{hyperref}
\else
\usepackage[pagebackref=true,breaklinks=true,colorlinks,bookmarks=false]{hyperref}
\fi

\ifwacvfinal
\else
\fi

\begin{document}

\title{An Ultra Lightweight CNN for Low Resource Circuit Component Recognition}

\author{Yingnan Ju, Yue Chen\\
Indiana University\\
{\tt\small \{yiju, yc59\}@iu.edu}
}

\maketitle

\begin{abstract}
   In this paper, we present an ultra lightweight system that can effectively recognize different circuit components in an image with very limited training data. Along with the system, we also release the data set we created for the task.  A two-stage approach is employed by our system. Selective search was applied to find the location of each circuit component. Based on its result, we crop the original image into smaller pieces. The pieces are then fed to the Convolutional Neural Network (CNN) for classification to identify each circuit component. It is of engineering significance and works well in circuit component recognition in a low resource setting. The accuracy of our system reaches 93.4\%, outperforming the support vector machine (SVM) baseline (75.00\%) and the existing state-of-the-art RetinaNet solutions (92.80\%).
\end{abstract}

\section{Introduction}

Efforts have been made to enable computers to automatically recognize different circuit or non-circuit components. Due to the development of mobile technology, sometimes the input can be as simple as a photo taken by a smart phone without any special lighting or background. The ideal output is the location and the information (name, description, confidence score, etc.) of each circuit component in the photo. This is a quick way to let students know circuit components by showing the names of various circuit components to them on their cellphone screen automatically and this can be applied to computer aided teaching circuit design and function in Engineering E101 class.
Our ultra lightweight system is developed to handle exactly this problem. It can be deployed on a server, a desktop computer, or even a smart phone. Our system architecture is simple and effective, relying less on complicated deep neural networks (DNN). 




To make it suitable for most use case scenario, we make every effort to simplify the design and make it a simple and effective system that can run on most laptops and smart phones. Some recent DNN-based solutions, like Faster R-CNN \cite{girshick2015fast} and RetinaNet \cite{lin2018focal} perform very well on some large data sets. However, in some specific scenario, like circuit component recognition with rather limited data and computing resources, our solution is the best performing and the most cost-effective.

\section{Related Work}

\subsubsection*{Selective Search}
Selective Search was first introduced to solve the problem of generating possible object locations for later use in object recognition \cite{van2011segmentation} \cite{uijlings2013selective}. Before that exhaustive search was the state-of-the-art \cite{van2011segmentation}. Selective search combines the strength of both exhaustive search and segmentation \cite{uijlings2013selective}. Instead of using one single technique to generate possible object locations and boxes, selective search uses a variety of complementary image partitioning to deal with as many image conditions as possible \cite{uijlings2013selective}. Usually, there are more than one circuit components in one photo, so selective search is the most suitable technique to divide the original photo into smaller images and each small image should contain only one circuit component.

\subsubsection*{Convolutional Neural Networks}
Current approaches to object recognition make use of machine learning methods \cite{krizhevsky2012imagenet}. In machine learning, a convolutional neural network is a class of deep, feed-forward artificial neural networks, and today such architectures are widely used for computer vision \cite{schmidhuber2015deep}. A large, deep convolutional neural network is capable of achieving high accuracy results on a highly challenging data set using purely supervised learning \cite{krizhevsky2012imagenet}. For a small data set, the result of classification from convolutional neural networks can sometimes be satisfying but the shortcomings of small data sets have also been widely recognized \cite{krizhevsky2012imagenet} \cite{pinto2008real}. 


\subsubsection*{SIFT}
Scale Invariant Feature Transform (SIFT) \cite{lowe2004distinctive}, is an image descriptor for image-based matching and recognition. This descriptor, as well as related image descriptors, are used for a large number of purposes in computer vision related to point matching between different views of a 3-D scene and view-based object recognition \cite{lindeberg2012scale}. In its original formulation, the SIFT descriptor comprised a method for detecting interest points from a grey-level image at which statistics of local gradient directions of image intensities were accumulated to give a summarizing description of the local image structures in a local neighborhood around each interest point, with the intention that this descriptor should be used for matching corresponding interest points between different images \cite{lindeberg2012scale}. If we get a library of images of all circuit components from the different angles and compare the photo newly taken with each image in the library, then the image in the library with most corresponding interest points with the newly taken photo should be the circuit component in that photo.

\subsubsection*{Faster R-CNN}
In 2014, S Ren et al.\cite{ren2015faster} proposed a method using selective search to extract regions from an image which is also known as region proposals \cite{girshick2014rich}. It takes a huge amount of time to train the neural network and about 47 seconds to segment each test image. In 2015, Ross Girshick et al. compensated some of the drawbacks of R-CNN and built a faster object detection algorithm called Fast R-CNN \cite{girshick2015fast}, which is significantly faster in training and testing sessions over the previous R-CNN. 
To bypass the problem of selecting a huge number of regions, S Ren et al.\cite{ren2015faster} proposed a method where they used selective search to extract just 2000 regions from the image. They called them region proposals. In the same year, a new object detection algorithm that eliminated the selective search algorithm called Faster R-CNN was proposed \cite{ren2015faster}. By 2017, Faster R-CNN, based on a two-stage approach popularized by R-CNN, had the highest accuracy for object detection \cite{lin2018focal}. 

\subsubsection*{RetinaNet}
The object detector with the highest accuracy by 2017 was the two-stage approach popularized by R-CNN, where a classifier was applied to a sparse set of candidate object locations \cite{lin2018focal}. Other deep neural network based one-stage algorithms, like YOLO or SDD had faster speed but lower accuracy \cite{redmon2016you}. The advantage of two-stage methods is that they first predict a few candidate object locations before using a convolutional neural network to classify each of them, while faster one-stage methods suffer from an extreme class imbalance. To counterbalance, RetinaNet proposed to reshape the standard cross entropy loss such that it down-weights the loss assigned to well-classified examples \cite{lin2018focal}. RetinaNet is able to match the speed of the previous one-stage detectors, like YOLO or SSD while surpassing the accuracy of all existing state-of-the-art two-stage detectors, like Faster R-CNN \cite{lin2018focal}.

\section{Methodology}

We divide our system into two main modules. The first module is for circuit component detection. The purpose of this module is to find the location and the size of the box of each circuit component in the image. Selective search is applied to detect various circuit components and find the most proper bounding box for each one. The second module is for circuit component recognition. The purpose of this module is to recognize the cropped image based on the candidate boxes from the result of the first module. A CNN is applied in this module to classify the cropped images. We also extract the features from the cropped images for our baseline SVM system.

\subsection{Data set for circuit components}

We release our circuit components recognition data set along with this paper. Our data set (total size: 2.25 GB) contains two subsets. One is a simple data set containing images of 13 different categories of circuit components (total size: 463 MB) and the other one is a full data set containing images of 30 different categories of circuit components (total size: 1.81 GB). Each subset contains raw images and categorized samples of cropped circuit components images and the details are listed below. We use the algorithms introduced in sub-section \ref{sec:iotcomponentdetection} and \ref{sec:iotcomponentrecognition} to crop the raw images into small pieces and each piece represents one circuit component. The cropped images are then annotated and categorized manually to build the part of categorized samples of cropped circuit components images. An extra category called "blank" is added for those false positive detection. 

Our data set is released under the Creative Commons Attribution-NonCommercial-ShareAlike 4.0 International License, CC BY-NA-SC 4.0 and can be obtained from one of the following ways: 

 \begin{enumerate}
   \item Onedrive sharing from an anonymous account:
   
\url{URL_annonymized}
   \item A hosted server:
   
\url{URL_annonymized}
 \end{enumerate}

\textit{a. Simple subset}

The simple data set are available as two separate zip files:

\begin{itemize}
\small
  \item Raw Images (raw\_images.zip, 421 MB): 262 images in JPEG format. 
     \begin{itemize}
     \item Camera model: Apple iPhone 6s
     \item Image Dimensions: $3024 \times 4032$
     \item Image Information: Bit depth - 24, Exposure Time - 1/30 sec, F-stop - f/2.2, DPI - 72, ISO speed - 50 \& 64
   \end{itemize}
  \item Categorized samples of cropped circuit components images (cropped\_samples.zip, 41.4 MB): 405 images in JPEG format. 
     \begin{itemize}
     \item Training set: 335 images in JPEG format in 14 folders.
     \item Test set: 70 images in JPEG format in 14 folders.
   \end{itemize}
\end{itemize}

13 categories of circuit components (with specific model): 

gy-521 module, 
ir receiver module, 
max7219 module, 
mega2560 controller board, 
pir motion sensor HC-SR501, 
power supply module, 
relay 5v, 
rotary encoder module, 
sound sensor module, 
stepper motor driver board uln2003, 
temperature and humidity module DHT11, 
ultrasonic sensor, 
water level detection sensor module.

\textit{b. Full subset}

The full data set are available as two separate zip files:

\begin{itemize}
\small
  \item Raw Images (raw\_images.zip, 1.73 GB): 1707 images in JPEG format. 
     \begin{itemize}
     \item Camera model: Apple iPhone 5s/6s/Air, Meilan Note, OnePlus 6
     \item Image Dimensions: $2448 \times 3264$, $4032 \times 3024$, $2592 \times 1936$, $1536 \times 2048$, $3104 \times 4192$, $3456 \times 4608$
     \item Image Information: Bit depth - 24, Exposure Time - 1/24 \& 1/30 \& 1/33 \& 50 sec, F-stop - f/1.7 \& f/2.2 \& f/2.4, DPI - 72, ISO speed - 40 \& 50 \& 64 \& 117 $\sim$ 202 \& 250
   \end{itemize}
  \item Categorized samples of cropped circuit components images (cropped\_samples.zip, 90.4 MB): 1757 images in JPEG format. 
     \begin{itemize}
     \item Training set: 1585 images in JPEG format in 31 folders.
     \item Test set: 172 images in JPEG format in 31 folders.
   \end{itemize}
\end{itemize}

30 categories of circuit components (with specific model): 

16 pin chip,
7 segment display,
9v battery,
buzzer,
capacitor,
ds3231 rtc module,
gy-521 module,
ir receiver module,
joystick module,
lcd module,
max7219 module,
mega2560 controller board,
membrance switch module,
motor,
pir motion sensor HC-SR501,
power supply module,
prototype expansion board,
relay 5v,
remote,
rfid module,
rotary encoder module,
servo motor,
sound sensor module,
stepper motor,
stepper motor driver board uln2003,
temperature and humidity module DHT11,
tilt ball switch,
transistor,
ultrasonic sensor,
water level detection sensor module.

\subsection{circuit component detection}
\label{sec:iotcomponentdetection}

Selective search is applied for circuit component detection with different diversification strategies. When selective search was firstly applied in object recognition by J.R.R. Uijlings et al. in \cite{uijlings2013selective} to deal with as many image conditions as possible, it was subject to following design considerations: capture all scales, diversification and fast to compute \cite{uijlings2013selective}. 

These considerations are similar with our goals in the system, but there is a major difference between our system and a general object recognition system. 


In a general object recognition system, a well-trained, generalized large deep neural network model is widely used to recognize objects after the objects are extracted from the image \cite{ren2015faster}\cite{krizhevsky2012imagenet}, like VGG \cite{simonyan2014very}, or Inception \cite{krizhevsky2012imagenet} model. These models have more than one thousand classes of images and were trained with 1.2 million images with another 50,000 images for validation \cite{simonyan2014very}\cite{russakovsky2015imagenet}.  That makes sure that these models have great capability to differentiate the objects extracted from the image with distinguishable confidence, and that is very helpful to filter the false positive result of selective research. In some systems, even after small similar regions of sub-segmentation were recursively combined into larger ones with greedy algorithm \cite{uijlings2013selective}, the result of selective search still includes many false positives and then these false positives will be filtered by the neural network if the confidence is lower than a threshold. However, in our system, there exists a small number of images manually taken for each circuit component to train the convolutional neural network model, and therefore the cropped image of false positive might also have high confidence ($>0.95$). That makes it difficult to set a threshold to filter those false positive results from selective search like in other systems. 


An important diversification strategy in selective search is to use different similarity measure $s_{ij}$ that captures the feature that affects the result \cite{van2011segmentation}\cite{uijlings2013selective}. The three similarity measures that we use were defined in \cite{van2011segmentation} and \cite{uijlings2013selective}:

\noindent
color similarity: 
\begin{equation}\label{equation:1}
s_{colour}(r_i,r_j)=\sum_{k=1}^n\min(c_i^k,c_j^k)
\end{equation}
size similarity: 
\begin{equation}\label{equation:2}
s_{size}(r_i,r_j)=1-\frac{size(r_i)+size(r_j)}{size(im)}
\end{equation}
shape compatibility:
\begin{equation}\label{equation:3}
fill(r_i,r_j)=1-\frac{size(BB_{ij})-size(r_i)-size(r_i)}{size(im)}
\end{equation}

The final similarity is a combination of the above three \cite{van2011segmentation}\cite{uijlings2013selective} where $r_i$ and $r_j$ are two regions or segments in the image and $a_1\in(0,1)$ denotes if the similarity measure is used or not:

\begin{equation}\label{equation:4}
s(r_i,\ r_j)=a_1s_{colour}(r_i,r_j)+a_2s_{size}(r_i,r_j)+a_3fill(r_i,r_j)
\end{equation}

To shorten the running time, we will try different combinations of the three similarities to see which ones are the most important for circuit component recognition on pure color background. In some experiments, we find out that even after the hierarchical grouping algorithm based on similarity, there are still some overlapped candidate boxes. Since the applied scenario is set to be easy, there should not be overlapped circuit components. Therefore, we apply some simple algorithm (Algorithm \ref{merging_algorithm}) to further group the overlapped boxes to make sure there is only one candidate box for each circuit component. The overlap rate between a box pair $(l_i,l_j)$ is defined by equation \ref{overlap_rate}. 

\begin{equation}
\label{overlap_rate}
Overlap(l_i,l_j)=\frac{area(l_i)\cap area(l_j)}{area(l_i)\cup area(l_j)}
\end{equation}

\begin{algorithm}
\label{merging_algorithm}
\SetAlgoLined
\KwResult{Grouped blocks}
 initialization\;
 set overlap = True\;
 \While{has overlap}{
  set overlap = False\;
  \ForEach {candidate box pair}{
  \If{overlap rate \textgreater THRESHOLD}{
  group the block pair to one box\;
  delete the original two boxes\;
  set overlap = True\;
  }
  }
 }
 \caption{Further candidate box grouping}
\end{algorithm}


Another diversification strategy is complementary color spaces \cite{van2011segmentation}\cite{uijlings2013selective}. Different color spaces have different invariants and different responses to changes in color \cite{uijlings2013selective}. We use RGB (red, green, blue) color space as the base line. We will also test HSV (hue, saturation, value) and LAB (lightness, a*, b*) color spaces. HSV and LAB color space are invariant to changes in bright and shadow.

To reduce the running time, the selective search was not applied directly on the original full-size image but on a reduced-size image with the same aspect ratio. The result of candidate boxes will be recalculated with a re-sizing index to crop small pieces of image of circuit components from the original image.

\subsection{circuit component recognition}
\label{sec:iotcomponentrecognition}

A CNN was applied for circuit component recognition. The original image of input was cropped into small pieces of image based on the candidate boxes of result of selective search and then these images were fed into the convolutional neural network to be classified to get the final result.

\begin{figure}
\centering
\includegraphics[width=0.5\textwidth]{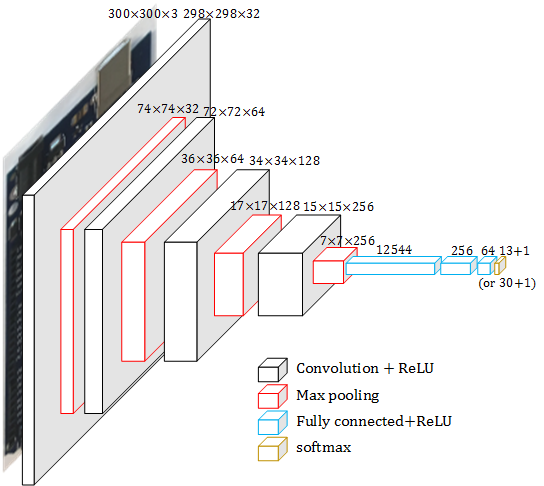}
\caption{Architecture of the convolutional neural network in the circuit component recognition part} 
\label{fig2}
\end{figure}

The architecture of the network in our system is shown in Fig \ref{fig2}. It contains seven layers with weights. The first four layers are convolutional layers and the rest three layers are fully connected layers. Max-pooling layers follow each convolutional layer. The output of the last fully connected layer is fed to a 14-way or 31-way (13 or 30 ways for 13 or 30 circuit components and 1 way for possible false positive background color) SoftMax covering these class labels for 30 kinds of circuit components in the experiment. Dropout is applied after the second and fourth layers to prevent overfitting of the model. The neurons in the last three fully connected layers are connected to all neurons in the previous layer. The ReLU non-linearity \cite{lecun2015deep} is applied to the output of every convolutional and fully connected layer (except the last fully-connected layer).

Take the $300\times300\times3$ input image as an example, the first convolutional layer filters the input image with 32 filters of $3\times3\times3$ with a stride of 1 pixel. The first pooling layer follows the first convolutional layer with window size $4\times4$ with a stride of 4 pixels. The output shape of the first pooling layer is $74\times74\times32$ and the second convolutional layer takes it as the input. The second convolutional layer has 64 filters of size $3\times3\times32$. The following pooling layer has $2\times2$ window size with a stride of 2 pixels. The output of the second pooling layer and also the input of the third convolutional layer has a shape of $36\times36\times64$. The third convolutional layer has 128 filters of size $3\times3\times128$ and the fourth convolutional layer has 256 filters of size $3\times3\times256$. The third and fourth pooling layers have the same $2\times2$ window size with a stride of 2 pixels with the second pooling layer. The output shape of the third and fourth pooling layers are $17\times17\times128$ and $7\times7\times256$ separately. The first two fully-connected layers have 256 and 64 neurons separately. The whole neural network has 3,617,294 parameters to train. For a network with $150\times150\times3$ input size, the parameters of the neural network can be reduced to 529,887.

The mini batch size of the convolutional neural network is set to 64 and the training is regularized by weight decay. The L2 penalty multiplier is set to $1\times{10}^{-6}$. The learning rate is set to $1\times{10}^{-6}$. The learning is stopped after 1000 epochs.

In our experiment, during the training part, the input of images are fixed-size images with three channels. Besides the $300\times300\times3$ input image mentioned in the example above, we also test $150\times150\times3$ and $200\times200\times3$ input images to find the balance between the accuracy of classification and the running time.

\subsection{SVM classification}
Support vector machines (SVM) were originally designed for binary classification \cite{hsu2002comparison}. Later researches \cite{weston1998multi}\cite{tsochantaridis2004support} effectively extended it for multi-class classification by combining several binary classifiers. Different from neural network accepting the raw or pixels of the original or re-sized image, we also try extract features from cropped images before we feed these features into SVM for classification. The feature we extract include aspect ratio of the image, color features and CenSurE (Center Surround Extremas) \cite{agrawal2008censure} features.

Color feature is important as the global features of an image \cite{lee2016style}. We extract the color features through statistical computation of the overall pixel of the cropped images. The color features include the average of hue and of saturation and the hue distribution. The average of hue and saturation can be calculated via equation \ref{average_hue} and \ref{average_sat}, in which M and N are dimension (width and height) of the image. The distribution of hue is extracted to show the ratio dispersion of each hue component in the image \cite{lee2016style} and can be calculated based on equation \ref{hue_distribution}. The CenSurE features are detected via a scale-invariant center-surround detector \cite{agrawal2008censure}. In our experiment, we extract CenSurE features and keep the largest n (n=10) values of CenSurE features as the input of SVM.

\begin{equation}
\label{average_hue}
f_{hue}=\frac{1}{MN}\sum_m\sum_n{}hue\quad
\end{equation}

\begin{equation}
\label{average_sat}
\quad f_{sat}=\frac{1}{MN}\sum_m\sum_n{}saturation
\end{equation}

\begin{equation}
\label{hue_distribution}
f_i=h_H(i)
\end{equation}


\section{Evaluation}

In this section we evaluate the quality of the selective search in circuit component detection part and the accuracy of convolutional neural network in circuit component recognition part. We will also test the accuracy and execution time of SVM and RetinaNet classification for comparison. For the selective search part, we use 20 raw photos from the simple subset of our data set; for other parts of evaluation, We use limited raw photos (less than 50 photos, ranging from 5M to 16M pixels) and the categorized cropped images from the full subset of our data set as the training set and test set and each photo contain some circuit components.

\subsection{Evaluation of circuit component detection}

In \cite{uijlings2013selective}, average best overlap (ABO) and mean average best overlap (MABO) were defined to evaluate the quality of object detection. The best overlap between each ground truth annotation $g_i^c\in G^c$ and the result of object detection L is calculated to get the average best overlap for one class c in \cite{uijlings2013selective}:
\begin{equation}
ABO=\frac{1}{|G^c|}\sum_{g_i^c\in G^c}\underset{l_j\in L}{\max}{Overlap(g_i^c,l_j)}
\end{equation}

The overlap score has the similar definition with equation \ref{overlap_rate} and it measures the area of the intersection of two regions divided by its union \cite{uijlings2013selective}. Mean average best overlap (MABO) was defined as the mean ABO over all classes \cite{uijlings2013selective}.

The first diversification strategy is complementary color spaces. The MABO and number of candidate boxes of different color spaces (RGB - baseline, HSV, LAB) is shown in the following Table \ref{table1}. The result of the number of box in the third column shows the boxes before algorithm \ref{merging_algorithm}.

\begin{table}
\centering
\caption{Mean average best overlap in different color spaces}
\label{table1}
\begin{tabular}{|l|l|l|}
\hline
Color spaces &  MABO & \#box\\
\hline
RGB & \textbf{87.96} & 20\\
HSV & 41.92 & 22\\
LAB & 81.95 & 197\\

\hline
\end{tabular}
\end{table}

From the Table 1 above, we observe large differences in results. We notice that RGB, as the baseline, has the best MABO score of 0.8975 using only 16 boxes for 13 categories of circuit components. HSV has the lower MABO score and LAB has good MABO score but uses much more boxes. Therefore, we will use RGB in the following experiments to find good combination of similarities.

The second diversification strategy is complementary similarity measures. We tried combination of the three similarities defined in (\ref{equation:1}), (\ref{equation:2}), (\ref{equation:3}) and (\ref{equation:4}) to see which ones has the best balance between MABO score, number of box and the running time. The result if shown in the following Table \ref{table2}.

\begin{table}
\centering
\caption{Mean average best overlap using different combination of similarities}
\label{table2}
\resizebox{\columnwidth}{!}{%
\begin{tabular}{|l|l|l|l|}
\hline
Similarities & MABO & \#box & Average running time\\
\hline
Color & 88.97 & 17 & 1.910s\\
Size & \textbf{89.76} & 20 & \textbf{1.875s}\\
Fill & \textbf{89.76} & 20 & 1.899s\\
Color+Size & \textbf{89.76} & 20 & 1.904s\\
Color+Fill & \textbf{89.76} & 20 & 1.903s\\
Size+Fill & \textbf{89.76} & 20 & 1.910s\\
Color+Size+Fill & \textbf{89.76} & 20 & 1.905s\\
\hline
\end{tabular}
}
\end{table}

From the Table \ref{table2} we see that single-color similarity has the lowest MABO score but it uses less boxes than other combinations do. The combinations of multiple similarities have slightly higher MABO score but use more boxes. The average running times for all combination are almost the same so it is not necessary for us to discard some similarity for a better running time.

\begin{table}
\centering
\caption{Mean average best overlap using different size of images}
\resizebox{\columnwidth}{!}{%
\label{table3}
\begin{tabular}{|l|l|l|l|}
\hline
Similarities & MABO & \#box & Average running time\\
\hline
$300\times 400$ & 89.76 & 20 & 1.905s\\
$600\times 800$ & 90.77 & 50 & 3.756s\\
$900\times 1200$ & 93.01 & 51 & 7.605s\\
$1200\times 1600$ & 91.95 & 51 & 13.998s\\
$1500\times 2000$ & 91.67 & 53 & 22.289s\\
$1800\times 2400$ & \textbf{93.10} & 50 & 36.686s\\
\hline
\end{tabular}
}
\end{table}

Table \ref{table3} shows the relation between running time (CPU: Intel E3-1245, GPU: Nvidia GTX 1070) and the size of the image. It can be seen from Table 3 that the running time of selective search increases significantly as the size of the image increase. Besides, although the images with higher resolution have slightly higher MABO score, the box they use and the average running time both increase significantly. Therefore, for most environments with limited computing hardware, we did not apply the selective search on the original full-size image but on a reduced-size image. However, it is also recommended to use larger images on much more powerful hardware for a better result.

\subsection{Evaluation of circuit component recognition}

The evaluation of circuit component recognition is shown in the following Table \ref{table5}.


\begin{table}
\centering
\caption{Accuracy of CNN with different input sizes}
\resizebox{\columnwidth}{!}{%
\label{table5}
\begin{tabular}{|l|l|l|l|}
\hline
Input size & circuit component & Final accuracy* & Testing time\\
~ & recognition accuracy & ~ & per image\\
\hline
$150\times150$ & 99.20 & 93.40 & 0.498ms\\
$200\times200$ & 99.20 & 93.40 & 0.561ms\\
$300\times300$ & \textbf{99.35} & \textbf{93.56} & 0.818ms\\
\hline
\end{tabular}
}
\begin{tablenotes}
    \item \footnotesize{* Final accuracy is influenced by both circuit component detection result and circuit component recognition accuracy.}
\end{tablenotes}
\raggedright{}
\end{table}

From the Table \ref{table5} we see that $150\times150$ size image and $300\times300$ size image as input have only slight difference in both performance and running time and the running time of CNN can be ignored compared with the running time of selective search. The difference between different input size only shows in the number of parameters of the neural network, size of model and the training time. The network with $150\times150\times3$ input size has 529,887 parameters and the size of the weights of the model is 2103 kB.

\begin{figure*}
\includegraphics[width=\textwidth]{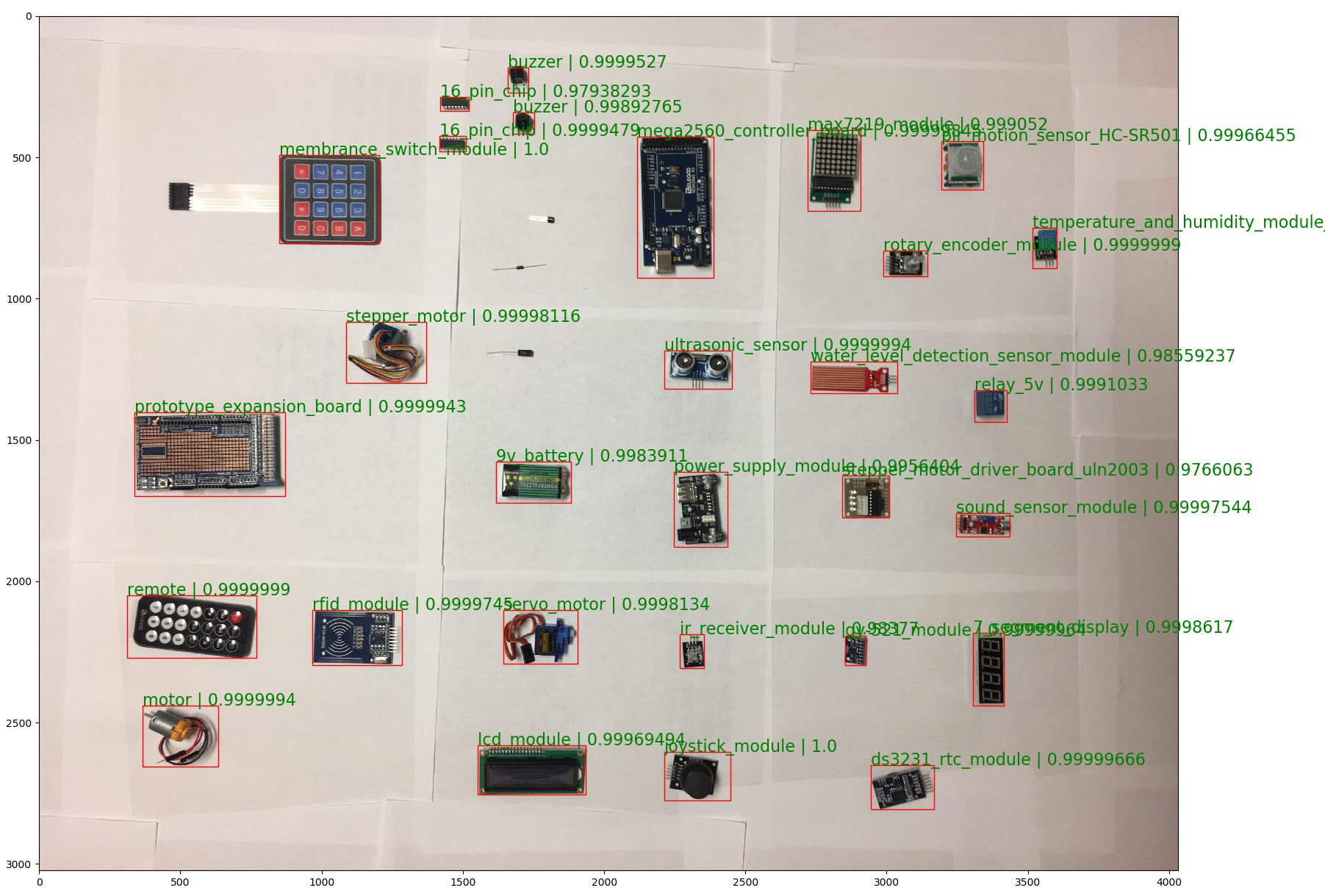}
\caption{Example of rendered images with the final result} 
\label{fig3}
\end{figure*}

The example of the final result of the system is shown as the follow Fig \ref{fig3}.

\subsection{Comparison}
To compare with our solution, we also experiment with SVM and Retinanet. For Retinanet, we apply the solution from \url{https://github.com/fizyr/keras-retinanet/} and use the annotated raw photo as the input.

We apply SVM classification and the result of SVM classification can be found in Table \ref{table6}. The input images are from the result of selective search. For feature extraction, the cropped images are re-sized to $150\times150$ pixels and the features are from equation (\ref{average_hue}) , (\ref{average_sat}) and (\ref{hue_distribution}). From the table we can see that the overall accuracy of classification of SVM is lower than the result of CNN, even after the optimization of SVD.

\begin{table}
\centering
\caption{Accuracy of SVM classification of cropped images}
\resizebox{\columnwidth}{!}{%
\label{table6}
\begin{tabular}{|l|l|l|}
\hline
Features & \#Features & Classification accuracy \\
\hline
Raw pixels & 22,500 & \textbf{80.60} \\
Aspect ratio + Average Hue & 3 & 68.70 \\
Color Features & 20 & 66.00 \\
CenSurE Features & 10 & 66.70 \\
\hline
\end{tabular}
}
\end{table}

The comparison between the best final result among our solution, SVM and RetinaNet can be found in Table \ref{table7}. From the comparison we can see that SVM has the lowest accuracy. The selective search + CNN solution and Retinanet solution both have good accuracy. The Retinanet is so far the state-of-the-art in object detection field and in our experiment has the fastest running time. However, it has much higher requirement to the annotation of the training set and much longer training time than the selective search + CNN solution.

\begin{table}
\centering
\caption{Comparison between three methods}
\resizebox{\columnwidth}{!}{%
\label{table7}
\begin{tabular}{|l|l|l|l|l|}
\hline
 & Classification & Final & Running \\
 & accuracy & accuracy & time \\
\hline
SVM* & 80.60 & 75.00 & 2.312s\\ 
Selective Search + CNN** & \textbf{99.20} & \textbf{93.40} & 1.981s \\ 
RetinaNet & - & 92.80 & 1.202s\\ 
\hline
\end{tabular}
}
\begin{tablenotes}
    \item \footnotesize{* The result of SVM is based on part circuit component (13/30).}
    \item \footnotesize{** The result of Selective Search + CNN is based on $150\times150$ input size.}
\end{tablenotes}
\raggedright{}
\end{table}

\section{Conclusion and future work}
In this paper, we apply selective search and CNN to implement a system that can recognize different circuit components in an image. 

Compared to traditional methods of classification, like SVM, the system of selective search + CNN is better both in terms of accuracy and running time. Compared to the state-of-the-art one-stage object detection/recognition solution - RetinaNet, our system has a slightly higher accuracy in our specific scenario. 

Another contribution is that We also release our circuit components recognition data set along with this paper. Our data set contains a simple data set containing images of 13 different categories of circuit components and another full data set containing images of 30 different categories of circuit components.The data set is released under the Creative Commons Attribution-NonCommercial-ShareAlike 4.0 International License, CC BY-NA-SC 4.0 and can be obtained by other researcher for free and open.




Although the best record of object recognition/detection is always broken by new solution nowadays, we are glad to see that an old method still has its day in some specific area. In actual engineering project, the final accuracy or the running time is not the only consideration and we need to balance the complexity and the availability of the system while not compromising to lower the accuracy. In the implementation, besides size of model and running time, dataset is our another consideration. In our experiment, we built the dataset from scratch. With our improved selective search algorithm, it is easy to crop the original photo into small pieces before we can manually put them into different categories. However, the format of training set of Retinanet is more complicated. It need the exact coordinate and the category of all circuit components in each photo. Actually, in our implementation, we used the result of selective search + CNN as the draft of the training set of Retinanet before we manually modified it to make it precise. That shows another advantage of our system when there is not enough training data. 

We do not deny the advantages of Retinanet and actually we envy its fast running time. If there is enough hardware and training data, Retinanet is the state-of-the-art solution. So far, it is not necessary to do circuit component recognition on mobile device in real time, so an about 2 seconds running time for each photo is not unacceptable, but in the next step, we are still going to try to shorten the running time of our system. If we take a deep look into the running time, we can see that the most part of time is wasted in selective search part. We have to use a thumb image to do the selective search but the time is still much longer than the time of CNN. If the processing time can be shortened more, the system might also be close to real-time and then it is also applicable to have an AR device as the client. Different from the cellphone, an AR device can display the information just beside each circuit component as a float layout in the field of view. That will be a much more direct and immersive experience for understanding different circuit components. 

For future experiments, we plan to shorten the running time of our system. If we take a deeper look into the running time, we can see that the most running time is used for selective search. If the processing time can be shortened, the system might be able to do real-time circuit component recognition.

\end{document}